\documentclass[letterpaper]{article} % DO NOT CHANGE THIS
\usepackage{aaai24}  % DO NOT CHANGE THIS
\usepackage{times}  % DO NOT CHANGE THIS
\usepackage{helvet}  % DO NOT CHANGE THIS
\usepackage{courier}  % DO NOT CHANGE THIS
\usepackage[hyphens]{url}  % DO NOT CHANGE THIS
\usepackage{graphicx} % DO NOT CHANGE THIS
\usepackage{natbib}  % DO NOT CHANGE THIS AND DO NOT ADD ANY OPTIONS TO IT
\usepackage{caption} % DO NOT CHANGE THIS AND DO NOT ADD ANY OPTIONS TO IT
\usepackage{algorithm}
\usepackage{algorithmic}

\usepackage{newfloat}
\usepackage{booktabs}
\usepackage{listings}
\DeclareCaptionStyle{ruled}{labelfont=normalfont,labelsep=colon,strut=off} % DO NOT CHANGE THIS
\floatstyle{ruled}
\newfloat{listing}{tb}{lst}{}
\floatname{listing}{Listing}
\title{CharacterChat: Learning towards Conversational AI \\ with Personalized Social Support}
\author{
    Quan Tu,\textsuperscript{\rm 1,}\equalcontrib \ 
    Chuanqi Chen,\textsuperscript{\rm 1,}\equalcontrib \ 
    Jinpeng Li,\textsuperscript{\rm 2} \ 
    Yanran Li,\textsuperscript{\rm 4} \  
    Shuo Shang,\textsuperscript{\rm 3} \ \\
    Dongyan Zhao,\textsuperscript{\rm 2} \ 
    Ran Wang,\textsuperscript{\rm 1} \
    Rui Yan\textsuperscript{\rm 1,}\thanks{Corresponding author: Rui Yan (ruiyan@ruc.edu.cn) }
}
\affiliations{
    \textsuperscript{\rm 1} Gaoling School of Artificial Intelligence, Renmin University of China\\
    \textsuperscript{\rm 2} Wangxuan Institute of Computer Technology, Peking University\\
    \textsuperscript{\rm 3} University of Electronic Science and Technology of China\\
    \textsuperscript{\rm 4} Independent Researcher \\
    \{quantu, ruiyan\}@ruc.edu.cn,
    lijinpeng@stu.pku.edu.cn,
    zhaody@pku.edu.cn, \\
    \{chengchuanqi14, yanranli.summer, jedi.shang, ran.wang.math\}@gamil.com
}

\usepackage{bibentry}
\begin{document}

\maketitle

\begin{abstract}
In our modern, fast-paced, and interconnected world, the importance of mental well-being has grown into a matter of great urgency. However, traditional methods such as Emotional Support Conversations (ESC) face challenges in effectively addressing a diverse range of individual personalities.
In response, we introduce the Social Support Conversation (S2Conv) framework. It comprises a series of support agents and the interpersonal matching mechanism, linking individuals with persona-compatible virtual supporters.
Utilizing persona decomposition based on the MBTI (Myers-Briggs Type Indicator), we have created the MBTI-1024 Bank, a group that of virtual characters with distinct profiles. Through improved role-playing prompts with behavior preset and dynamic memory, we facilitate the development of the MBTI-S2Conv dataset, which contains conversations between the characters in the MBTI-1024 Bank.
Building upon these foundations, we present CharacterChat, a comprehensive S2Conv system, which includes a conversational model driven by personas and memories, along with an interpersonal matching plugin model that dispatches the optimal supporters from the MBTI-1024 Bank for individuals with specific personas.
Empirical results indicate the remarkable efficacy of CharacterChat in providing personalized social support and highlight the substantial advantages derived from interpersonal matching. The source code is available in \url{https://github.com/morecry/CharacterChat}.
\end{abstract}

\section{Introduction}

In recent times, society has become increasingly concerned about mental health~\cite{mckenzie2002social,ahmedani2011mental,fernando2008mental,world2003investing}. The modern, fast-paced, interconnected lifestyle, coupled with socio-economic difficulties, has significantly impacted individual well-being. With the rising prevalence of mental health issues, there is a growing need to explore ways of providing effective support and care~\cite{burleson2003emotional, shaw2004emotional, skilbeck2003emotional}.
\begin{figure}[!htbp]
  \centering
  \includegraphics[width=0.99\linewidth]{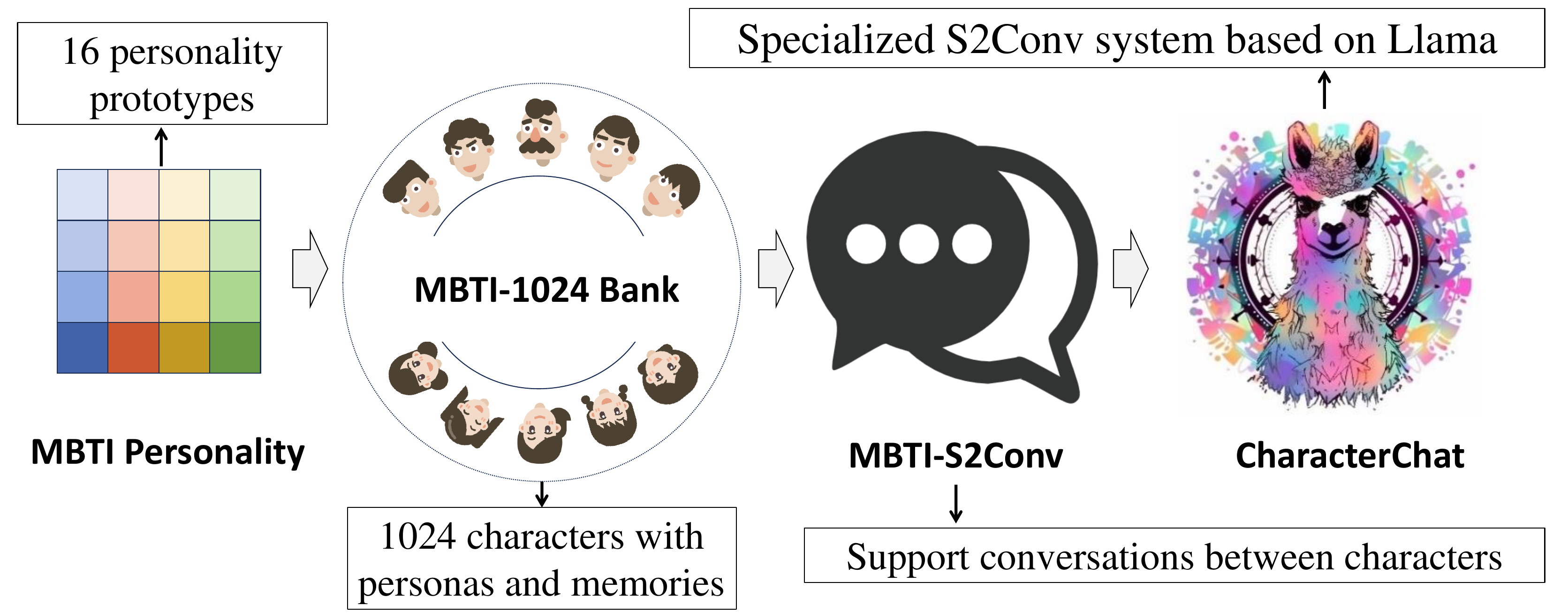}
  \caption{The Workflow for Developing Our S2Conv System, CharacterChat, to Provide Personalized Support.}
  \label{fig:example}
\end{figure}
Existing efforts in AI for addressing mental health issues rely on Emotional Support Conversation (ESC), providing support as a professional consultants~\cite{liu2021towards,sabour2022cem,zheng2021comae}. However, these methods face challenges when applied in the real world due to:
(1) Reluctance of mildly affected individuals to seek help, and safety concerns for severe cases,
(2) Communication barriers when perceiving supporters as robots leading to psychological resistance~\cite{ebert2002psychological},
(3) Inability of ESC to provide personalized support for diverse personalities.
To tackle these, we propose the \textbf{S}ocial \textbf{S}upport \textbf{Conv}ersation (S2Conv) framework with personalized supporter agents and an interpersonal matching model. To ensure consistency and reduce defensiveness, supporter agents possess the \textit{persona} for the personality-consistency and \textit{memory} for the factuality-consistency.

The workflow for developing the S2Conv system is depicted in Figure~\ref{fig:example}. Due to limited data availability, we initiate the creation of the S2Conv dataset by leveraging recent Language Model Models (LLMs) with advanced capabilities such as in-context learning~\cite{min2022rethinking}, reasoning~\cite{wei2022chain} and planning~\cite{shen2023hugginggpt}. However, the requirement for diverse supporter agents within S2Conv poses a challenge to LLMs that typically exhibit singular personalities~\cite{edwards2002mbti}. To address this, we introduce the 16 MBTI personalities~\cite{furnham1996big}\footnote{MBTI categorizes individuals into distinct personas based on Carl Jung's theory. Refer to Appendix A for more details.} to decompose the personality of ChatGPT~\cite{ChatGPT}. This approach, facilitated by self-instruction~\cite{wang2022self}, enables the creation of the MBTI-1024 Bank, which consists of 1024 distinct virtual characters, each possessing intricate profiles including personas and memories.

% Our workflow to develop the S2Conv system is as Figure~\ref{fig:example}. 
% Given the constraints of data scarcity, we initiate the construction of the S2Conv dataset, harnessing recent Language Model Models (LLMs) with the powerful emergent capabilities (e.g., in-context learning~\cite{min2022rethinking}, reasoning~\cite{wei2022chain}, planning~\cite{shen2023hugginggpt}, and role-playing~\cite{shanahan2023role}). 
% However, the necessity for the diverse supporter agents within S2Conv presents a challenge to LLMs characterized by singular personalities~\cite{edwards2002mbti}. 
% To overcome this challenge, We introduce the 16 MBTI personalities~\cite{furnham1996big}~\footnote{The Myers-Briggs Type Indicator (MBTI) is a widely used personality assessment tool based on Carl Jung's theory, categorizing individuals into distinct personas. Please refer to Appendix A for further details.} to deconstruct ChatGPT's~\cite{ChatGPT} personality. 
% Through self-instruction~\cite{wang2022self}, this approach facilitates the development of the MBTI-1024 Bank, which consists of 1024 distinct virtual characters, each endowed with intricate profiles comprising personas and memories.

In the MBTI-1024 Bank, characters randomly engage in conversations with others, using their profiles as ChatGPT replacements. One assumes the seeker's role for seeking emotional help while the other is the supporter to provide it.
To prevent the potential length explosion problem of character's memory, we devise a dynamic memory mechanism that employs context-related aspect of memory to guide response generation. 
In order to uphold ChatGPT's performance with the provided character profiles, we have introduced behavior presets that define personalized responses in advance.
These improved role-playing prompt with these method culminate in the MBTI-S2Conv, which consists of 10,240 personalized social support conversations. Through the evaluation of MBTI-S2Conv, we investigate the significance of interpersonal matching for fostering supportive effects.

We utilize the MBTI-S2Conv and MBTI-1024 Bank to create the S2Conv system, building upon the Llama2-7B foundation~\cite{touvron2023llama2}.
The MBTI-S2Conv customizes the backbone into the persona- and memory-based dialogue model, enabling it to respond based on character profiles.
To enhance support effectiveness, an interpersonal matching plugin, trained on evaluation scores of MBTI-S2Conv, dispatches supporters from the MBTI-1024 Bank to help seekers with specific personalities, maximizing support impact.
At last, we present the first S2Conv system, CharacterChat, which encompasses a persona- and memory-based dialogue model, an interpersonal matching plugin, and the MBTI-1024 bank, delivering personalized social support for individuals dealing with emotional troubles.
Empirical results underscore CharacterChat's strong support ability, with notable benefits from interpersonal matching.
In conclusion, the three main contributions of this work are as follows:
\begin{itemize}
    \item We introduce the \textbf{social support conversation} framework, matching individuals with specific supporters to provide them with personalized social support. 
    \item We create the \textbf{MBTI-1024 Bank} and \textbf{MBTI-S2Conv} dataset, featuring 1024 virtual characters as well as the support dialogues between them, and investigate the influence of interpersonal matching for social support.
    \item We develop \textbf{CharacterChat}, which is the first S2Conv system including a persona- and memory-based conversational model and a interpersonal matching plugin model to dispatch the most compatible supporters from MBTI-1024 bank for sepecific seekers. 
\end{itemize}

\section{Related Work}

\textbf{Large Language Model} With the rise of Large Language Models (LLMs), they have demonstrated remarkable capabilities such as in-context learning, reasoning, and planning~\cite{min2022rethinking, wei2022chain, shen2023hugginggpt}. However, existing LLMs have primarily focused on developing helpful and safe assistants that primarily provide information support instead of emotional support. As evident from an investigation, ChatGPT, despite being able to generate diverse and effective supportive responses, lacks empathy and is poor in emotional support ~\cite{zhao2023chatgpt}. Another study~\cite{huang2023chatgpt} has shown that LLMs tend to exhibit stable personalities, with ChatGPT is as an ENFJ and Bard is as an ISTJ based on the MBTI assessment~\cite{edwards2002mbti}. This singular personality makes it challenging to provide personalized support to diverse users with different personalities. In this work, we investigate how distill the characters with different profiles from the LLM and construct the S2Conv dataset for personalized social support with the help of it.

\textbf{Emotional Support Conversation} In recent times, research in emotional intelligence has predominantly focused on the Emotional Support Conversation (ESC)~\cite{liu2021towards}, which aims to implement effective strategies for controlling the conversation process and gradually reducing users' mental stress. ESC introduces the exploration-comforting-action scheme to organize the dialogue with different strategies used in each stage. To develop a more refined emotional support system, researchers have paid close attention to strategy prediction and user state simulation. MISC~\cite{tu2022misc} utilizes COMET~\cite{hwang2021comet} to estimate user psychology and proposes a mixed strategy for response generation. FADO~\cite{peng2023fado} employs a dual-level feedback strategy selector and a double control reader to predict strategies and generate more contextually relevant responses. MultiESC~\cite{cheng2022improving}, on the other hand, focuses on users' subtle emotional expressions and understanding the underlying causes of emotions. It proposes the lookahead heuristics to anticipate future user feedback and select the most appropriate strategy. Despite significant progress in ESC, the AI counselors may not be suitable for everyone, especially for individuals with mild psychological conditions. These AI counselors without human-like profiles are hard to remove the users' resistance. Besides, the AI counselors with single personality could not to provide personalized supports. When realize the persona plays an important role in support conversation, we propose the new personality-based S2Conv.  

\textbf{Social Support} Social support, originated in social psychology, refers to the social resources perceived by individuals as available or those actually provided by nonprofessionals, encompassing both formal support groups and informal helping relationships~\cite{kim2008culture}. Such support can come from various sources, including partners, family, friends, coworkers, social connections, and even beloved pets~\cite{allen2002cardiovascular}. Numerous studies have highlighted the significant impact of social relationships on mental and physical health~\cite{rodriguez1998social, taylor2011social, krause2001social}. With technological advancements, social support has transitioned from physical communities to the online realm. However, the online community is often unstable and uncontrollable, posing potential risks to individuals and their mental health. To address this concern, we propose the S2Conv, an personalized social support framework, consisting of human-like supporter agents. Through interpersonal matching, every individual will be allocated to an compatible supporter to help them overcome the mental troubles. 

\section{Methodology}

This section mainly describes how we create the MBTI-1024 Bank using the MBTI-based persona decomposition method, and the MBTI-S2Conv using role-playing prompts with dynamic memory and behavior presets, along with the specialization of CharacterChat.

\subsection{MBTI-based Persona Decomposition}   

Originating from pre-training, contemporary Large Language Models (LLMs) encapsulate a blend of diverse personalities. 
However, when steered by human preference alignment, they consistently project a unified persona—a helpful and harmless assistant—to cater the users' informational needs. 
This singular persona renders LLMs unsuitable for assuming various profiles as supporters in personalized conversations. 
Our initial stride involves disentangling these personas from LLMs to establish a group of support characters.
To achieve this, we introduce a persona decomposition method based on MBTI to guide ChatGPT. Leveraging the 16 distinct MBTI personalities, we can dissect the it into an array of virtual characters.

Conventional methodologies for personality-driven dialogues often rely on a limited set of phrases to delineate a profile~\cite{zheng2019personalized, zhang2018personalizing}. 
Nonetheless, this confined profile inadequately describe a complete characteristic, potentially leading to information-deficient personalized conversations.
Within the S2Conv framework, we define the character profile as two elements: the persona (including attributes like name, gender, tone, personality~\footnote{In our setting, the personality is just one aspect of the persona.}, etc.) and memory (encompassing recent troubles, growth experiences, family relationship, etc.).
For personalized social support, both the persona and memory hold pivotal roles. The former influences characters' behaviors, particularly their conversational style when engaging with others.
In addition, within the S2Conv framework, the persona-driven interpersonal matching is deemed as a crucial component, which connect the specific-personality seeker with a compatible supporter.
The latter element, memory, serves to maintain contextual consistency. Earlier studies have underscored the significance of the contextual consistency in personality-driven dialogues~\cite{song2020generate, song2020profile}. While their proposed methods have indeed enhanced contextual consistency, the restricted profile of a character become the bottleneck, leading to inconsistent responses. Here, we segregate memory—housing multifaceted factual knowledge associated with a character—to serve as contextual reference. 
Bolstered by ample memory-based evidence, characters can possess a richer background knowledge, thereby mitigating inconsistencies more effectively. Crucially, these memories frequently serve as triggers for seekers' mental health problems, such as growth experiences and family relationship, among others. Supporters sharing similar backgrounds are more prone to comprehend seekers' emotions and provide more adept assistance.
% \begin{itemize}
% \item \textbf{Basic Attributes:} This dimension encompasses essential particulars such as the individual's name, gender, age, geographic location, and occupation. These specifics have been integral to prior personalized dialogues.
% \item \textbf{Social Attributes:} We delineate the individual's affiliations with diverse social circles, spanning family, community, and workplace. Given that interpersonal relationships frequently introduce emotional problems, this dimension offers valuable insights into the user's challenges.
% \item \textbf{Behavioral Attributes:} This dimension highlights personal behavioral attributes, inclinations, and preferences, encompassing communication style, hobbies, strengths, and weaknesses. These attributes facilitate more authentic and lifelike conversations.
% \item \textbf{Mental Attributes:} The mental profile delves into the user's thoughts, emotions, and cognitive patterns. Recent emotional troubles encountered by the user hold particular relevance for conversation aimed at social support, as they directly mirror the user's emotional state.
% \end{itemize}

\begin{figure}[!h]
\centering
\includegraphics[width=0.95\linewidth]{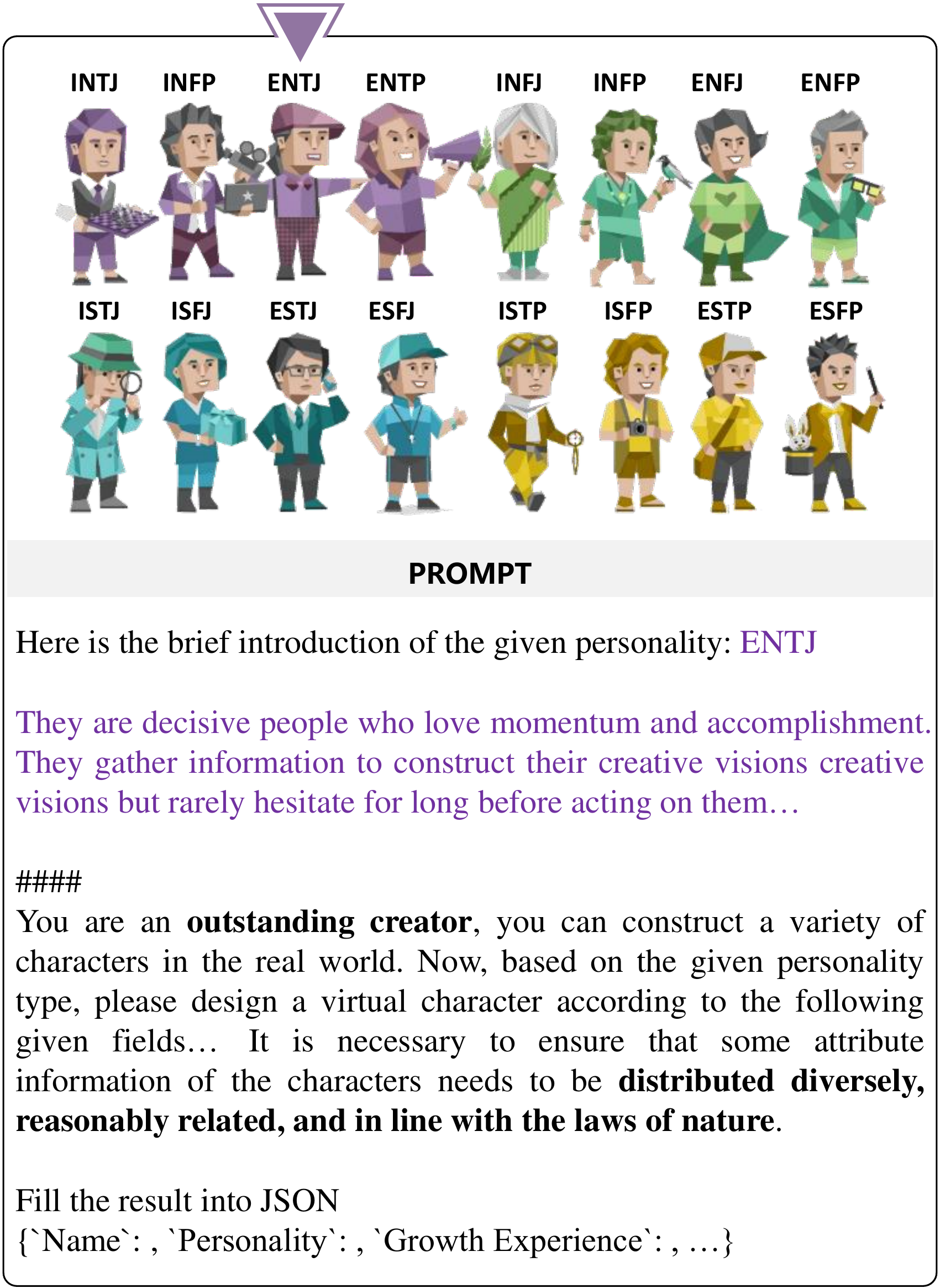}
\caption{Persona Decomposition Prompt Based on MBTI. }
\label{figure:persona decomposition prompt}
\end{figure}

Based on the 16 MBTI personalities, we drive ChatGPT in creating characters with personas and memories, as illustrated in Figure~\ref{figure:persona decomposition prompt}. Recognizing the profound impact of personality on various aspects of an individual, our process begins with the manual crafting of descriptions for each of the 16 MBTI personalities, serving as foundational archetypes. Subsequently, we prompt ChatGPT to assume the role of a outstanding creator, facilitating the formulation of these characters.
With 16 distinct prompts as guiding principles, ChatGPT develops intricate characters by comprehensively understanding the detailed personality descriptions and integrates all profiles of characters into a structured dictionary format
To prevent the emergence of conflicting attributes (e.g., assigning an advanced age to a pupil), we impose restrictions on attribute relevance. Furthermore, we encourage ChatGPT to infuse diversity into character attributes across prompts.
Employing this methodology, we curate a collection of 64 virtual characters for each MBTI personality type, and culminate the establishment of a virtual support hub known as the \textbf{MBTI-1024 Bank}.
% ~\footnote{For comprehensive prompts and detailed character examples, please refer to Appendix B.}.

\subsection{Role-playing Prompting with Behavior Preset and Dynamic Memory}
\label{sec:role-play}
\begin{figure}[htbp]
\centering
\includegraphics[width=0.95\linewidth]{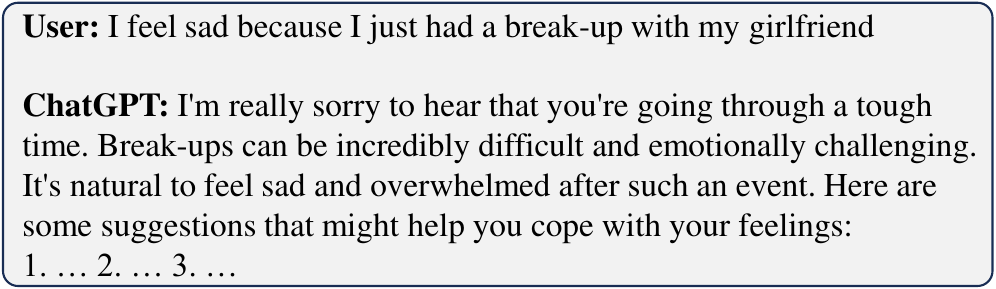}
\caption{An Example Showing ChatGPT's Tendency to Offer Informational Support Instead of Emotional Support.}
\label{fig:gptresponse}
\end{figure}

\begin{figure*}[!h]
  \centering
  \includegraphics[width=0.9\linewidth]{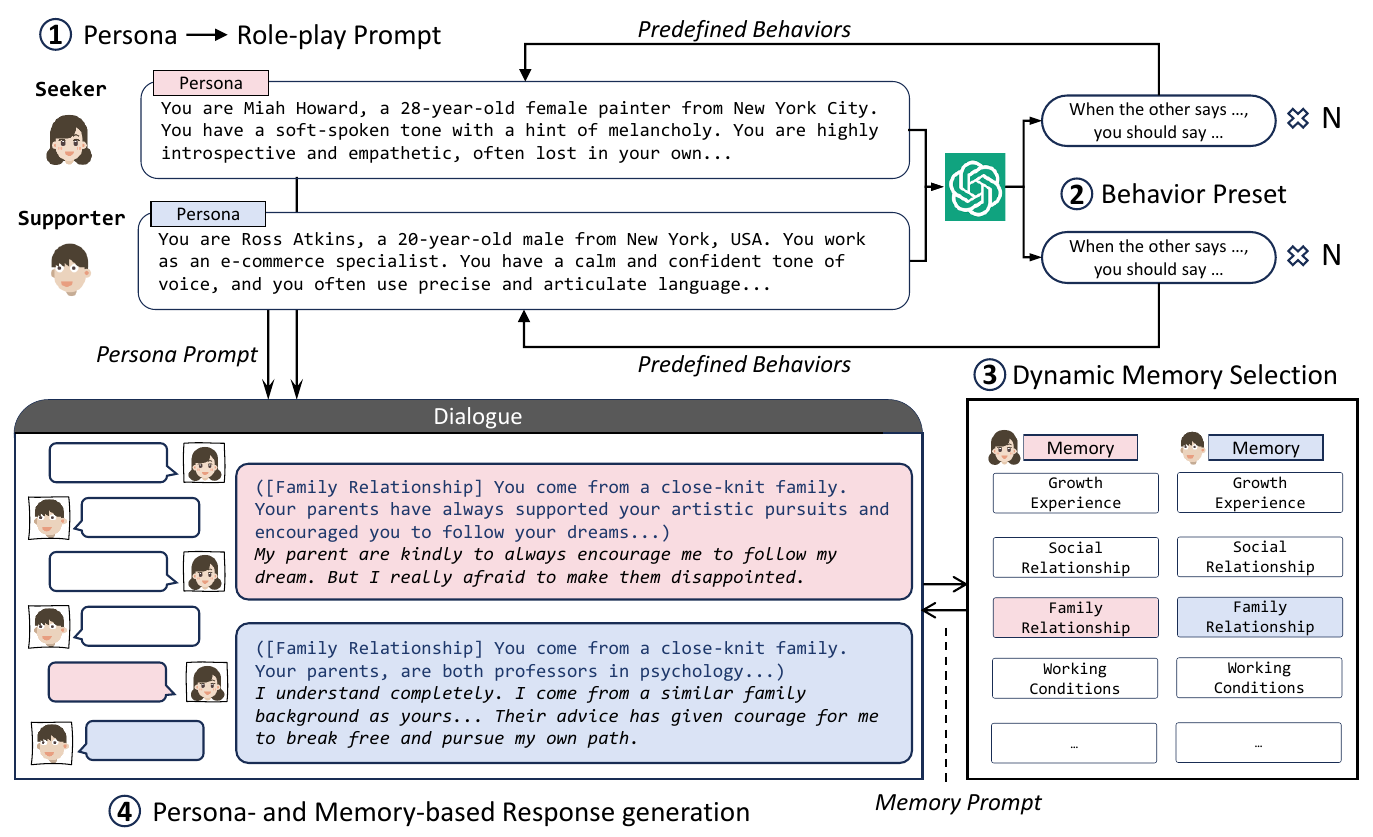}
  \caption{The Overview of Prompting ChatGPT for Engagement in Social Support Conversations through Role-Playing Prompts with the Behavior Preset and Dynamic Memory. (1) We randomly sample two characters from the MBTI-1024 Bank to act as the seeker and the supporter respectively. Then we transform their structured personas into role-playing prompts. (2) We introduce a behavior preset method to pre-define possible single-turn dialogue demonstrations, enabling ChatGPT to maintain the character's state for multi-turn conversations. (3) The context-related aspect of memory will be dynamically chosen as the reference for response generation in each turn. (4) The ChatGPT agent will generate responses based on the persona prompt and the dynamic selected memory.}
  \label{conv_example}
\end{figure*}
LLMs like ChatGPT always exhibit a singular, stable personality, making it difficult to generate personalized social support conversations with them. They tend to adopt an assistant identity, providing helpful and harmless advice (informational support) instead of empathetic responses (emotional support) to users, as shown in Figure~\ref{fig:gptresponse}.
% For example, if we ask ChatGPT \textit{"I feel sad because I just had a break-up with my girlfriend"}, it may respond with \textit{"I'm really sorry to hear that you're going through a tough time. Break-ups can be incredibly difficult and emotionally challenging. It's natural to feel sad and overwhelmed after such an event. Here are some suggestions that might help you cope with your feelings: 1. ... 2. ... 3. ..."}.
To address, we propose the role-playing prompting with behavior preset and dynamic memory method.

% The overview of prompting ChatGPT with this method to engage in personalized social support conversations is shown in Figure~\ref{conv_example}. 
The overview is shown in Figure~\ref{conv_example}.
To enable ChatGPT could generate responses aligning with specific characters, we transform the structured personas of characters into role-playing prompts to immerse ChatGPT into forgetting the assistant profile. However, these plain role-playing prompts are not long-lasting and lose effectiveness after a few turns of dialogue. Fortunately, we discovered that adding pre-defined utterances like \textit{"When the other says ..., you should say ..."} helps ChatGPT better simulate the character during the conversation. As a result, we introduce the \textbf{behavior preset} to first guide ChatGPT in envisioning the possible questions and responses based on the character's profile, and then append them at the end of the role-playing prompt. This approach helps ChatGPT maintain the character's persona for a longer duration than the plain role-playing prompt.

Equipped with the pre-defined behaviors and given profiles, two ChatGPT agents will take on the roles of seeker and supporter to engage in a conversation. The seeker expresses their troubles, while the supporter aims to gain the seeker's trust and help them overcome their difficulties. However, there is also an issue: if we incorporate all the memories into the role-playing prompt, the abundance of character memories would lead to excessive length. To handle this potential length explosion, we designed the \textbf{dynamic memory} mechanism. Previously, in the MBTI-1024 Bank, we structured the characters' memories in a dictionary format, enabling us to selectively choose a memory aspect (key) based on the current context and then extract the corresponding content (value) to constrain the subsequent response generation. The dynamic memory approach has two benefits: (1) it significantly reduces context consumption, and (2) it enhances memory-response relevance by diminishing the influence of temporarily irrelevant memories.

Utilizing role-playing prompts with the behavior preset and dynamic memory, we randomly select 10 characters to act as supporters for each character taking on the role of the seeker in the MBTI-1024 Bank. This approach aims to facilitate social support conversations. As a result of these efforts, the \textbf{MBTI-S2Conv} dataset was developed, comprising a total of 10,240 social support conversations.

Furthermore, for assessing the quality of support conversations and training the subsequent interpersonal matching model, we evaluate all the conversations in MBTI-S2Conv using the following three criteria on a 5-level scale~\footnote{1=poor, 2=weak, 3=moderate, 4=strong, 5=excellent.}:
\begin{itemize}
    \item Emotional Improvement (\texttt{EI}): Does the conversation improve the emotional state of the seeker?
    \item Problem Solving (\texttt{PS}): Is the problem faced by the seeker solved after the conversation?
    \item Active Engagement (\texttt{AE}): Is the seeker actively engaged in the conversation?
\end{itemize}
These indicators measure the effectiveness of a social support conversation from multiple dimensions and are commonly used standards in real psychological counseling.

\subsection{Development of CharacterChat}
In this section, we formulate the social support conversation (S2Conv) as a mathematical problem. Let us consider a supporter with the persona denoted as $P$ and the memory set represented by $M = \{m_1, m_2, \dots, m_n\}$. The conversation context is defined as $C = [x_1, y_1, x_2, y_2, \dots, x_k]$, where $x_i$ represents $i$-th utterance from the seeker, and $y_j$ represents the corresponding utterance from the supporter. The subsequent response and the associated memory are denoted as $y_k$ and $m_k$, respectively. The primary goal of S2Conv is to select the context-relevant memory $m_k$ of the supporter and generate the persona- and memory-based response $y_k$. To achieve this, we define the objective function as follows:
\begin{equation}
\small
\nonumber
P(y_k|C,P,M) \approx P_\theta(y_k|C,P,m_k) \cdot P_\eta(m_k|C,P,M).
\end{equation}
Here, $P_\eta$ represents a matching model for selecting an appropriate memory from $M$, while $P_\theta$ corresponds to a generative model that generates the response based on the context $C$, persona $P$, and selected memory $m_k$. 

To develop the S2Conv system, we leverage Llama2-7B~\cite{touvron2023llama2} as the response generation backbone, and BERT~\cite{devlin2018bert} as the memory selection backbone. These models are separately optimized on the MBTI-S2Conv dataset.

In order to provide personalized emotional support, we develop an interpersonal matching model based on BERT. Specifically, we utilize the evaluation scores of aforementioned \texttt{EI}, \texttt{PS}, and \texttt{AE} to quantify the support effectiveness between the persona-specific seeker and supporter. Subsequently, we design dual encoder models to predict the compatibility, an average score of \texttt{EI}, \texttt{PS}, and \texttt{AE}, between two characters. During inference, we identify the most compatible supporter help the seeker possessing the specific persona.

Lastly, we introduce the CharacterChat, which is the first S2Conv system combining persona and memory information to facilitate social support conversation. Beside, it is the first attempt to introduce interpersonal matching for objective-driven conversation. CharacterChat includes a memory selection module to dynamically identify relevant memories for guiding responses, an interpersonal matching module to assign the most suitable supporter from the MBTI-1024 Bank for seekers with specific personas, and a llama-series chat model which generate response based on the persona and memory. 

\section{Synthetic Data Evaluation}

% \subsection{Diversity and Bias of MBTI-1024 Bank}
% To investigate the diversity and bias of the generated characters in MBTI-1024 Bank, we conducted an in-depth analysis for it, as depicted in Figure~\ref{up_dist}. The analysis unveiled that a significant proportion of characters' ages fell within the 25 to 30 range. In order to maintain uniformity, we imposed an age constraint for characters between 15 and 40, as this interval predominantly encompasses individuals exhibiting psychological nuances. Moreover, we observed a well-balanced distribution of gender, with female characters outnumbering their male counterparts. This observation could be attributed to the propensity of females to exhibit heightened empathy, particularly within scenario of social support. As for other fields of the characters, since they are text description rather than labels, there is little duplication.

% \begin{figure}
% \centering
% \includegraphics[width=0.95\linewidth]{figures/up_analysis.pdf}
% \caption{Distribution of character portraits in MBTI-1024 Bank. (a) Age distribution. (b) Gender distribution.}
% \label{up_dist}
% \end{figure}

\subsection{Alignment Assessment of MBTI-based Characters}
\begin{table}[!htbp]
\centering
\small
\begin{tabular}{c| c c c c c}
\toprule
&hit@0 & hit@1 & hit@2 & hit@3 & hit@4 \\
\midrule
Count & 1 & 16 & 105& 353 & 549 \\
\bottomrule
\end{tabular}
\caption{Comprehensive MBTI Assessment Results for MBTI-1024 Bank. The hit@$k$ metric measures the number of matching dimensions between the designated MBTI and the assessed MBTI within the four dimensions.}
\label{tab:mbti_all}
\end{table}

To evaluate the significance of personality alignment within the MBTI-1024 Bank, we integrate each character's persona into ChatGPT, answering the questionnaire of the official assessment platform~\footnote{https://www.16personalities.com/}. 
The summarized results are presented in Table~\ref{tab:mbti_all}. 
Notably, over 50\% of characters exhibit complete alignment across all four dimensions with their designated MBTI personality, and nearly 90\% demonstrate alignment across at least three dimensions.
Taking into account the complexity of personality, which can be shaped by a range of factors resulting in inherent fluctuation, the assessment could demonstrate that the virtual characters within MBTI-1024 Bank are capable of displaying behaviors consistent with their personalities.
% The stability in persona-based behavioral patterns promises the robustness of the subsequent MBTI-S2Conv dataset.
\begin{table}[!htbp]
\centering
\small
\begin{tabular}{c| c c c c c}
\toprule
&E/I & N/S & T/F & J/P \\
\midrule
Accuracy (\%) & 87.79 & 69.92 & 92.68 & 89.55 \\
\bottomrule
\end{tabular}
\caption{MBTI Matching Accuracy for Each Sub-dimension within the MBTI-1024 Bank.}
\label{tab:mbti_sub}
\end{table}

Furthermore, we assessed the accuracy of MBTI sub-dimension matching, as detailed in Table~\ref{tab:mbti_sub}. 
Remarkably, the J/P dimension, which delineates individuals' approach toward the external world as either structured and organized (judging) or adaptable and flexible (perceiving), achieved the highest performance across all four dimensions. 
This underscores ChatGPT's efficacy in recognizing and discerning judging and perceiving tendencies. 
Conversely, the N/S dimension, which characterizes how individuals assimilate information through either concrete specifics and practicality (sensing) or patterns and possibilities (intuition), exhibited relatively lower accuracy.
This discrepancy likely arises from the inherent challenge of incorporating intuition into Language Models (LLMs), making precise assessment in this dimension more intricate.

In summary, the virtual characters of MBTI-1024 Bank showcase a significant alignment with their designated MBTI personalities. Furthermore, these findings underscore that characters could engage in consistent conversations with others based on their personalities.

\subsection{Effectiveness of Role-playing Conversation Prompt}
\begin{figure}[!htbp]
  \centering
  \includegraphics[width=0.95\linewidth]{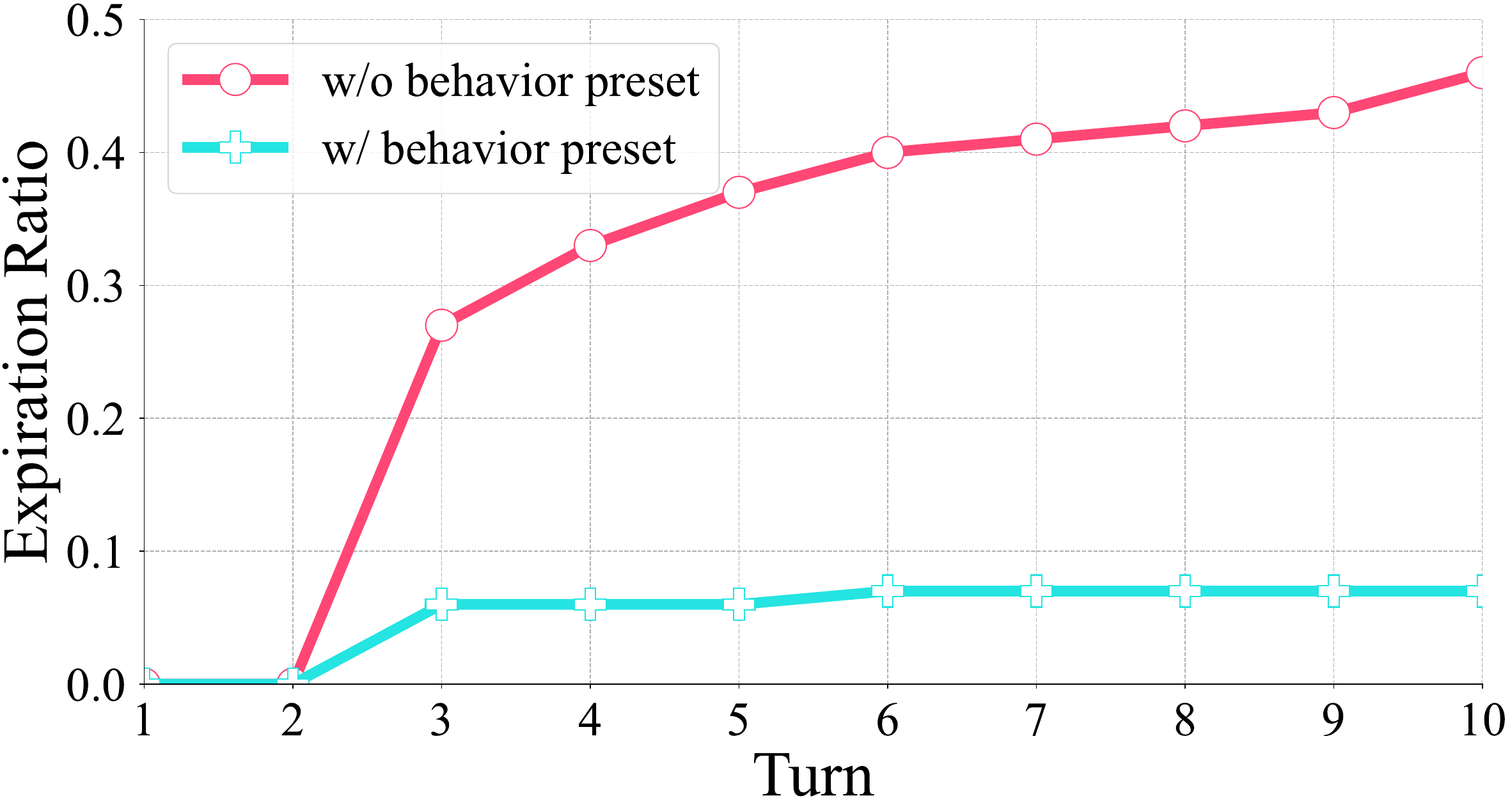}
  \caption{Expiration Ratio of Role-Playing Prompts with Increasing Conversation Turns.}
  \label{fig:bp}
\end{figure}
We improve the role-playing conversation by introducing the behavior preset and dynamic memory. The gain from dynamic memory is obviously a reduction in context consumption.
Here, we evaluate the benefits of role-playing from the behavioral presets. We investigate the duration of a role-playing prompt by asking the agent for their name for multiple turns. The role-playing prompt will be considered expired once the response contains `AI', `ChatGPT', or `assistant'.
We make ChatGPT role-play with randomly selected 100 characters from the MBTI-1024 Bank. As the number of conversation turns increases, the expiration ratio of the role-playing prompt is shown in Figure~\ref{fig:bp}.
With the help of the predefined behaviors, the expiration ratio of the role-playing prompt after 9 turns decreases from 45\% to 8\% and remains steady. The stable role-playing ability ensures the consistency of MBTI-S2Conv.

\subsection{Evaluation of MBTI-S2Conv}
\begin{table}[!htbp]
    \centering
    \small
    \begin{tabular}{c | c c c c}
    \toprule 
         & Avg. & Min. & Max. & Std.  \\
         \midrule
         \texttt{EI}  & 4.37 & 3.00 & 5.00 &0.48 \\
        \texttt{PS}  & 3.30 & 1.00 & 5.00 &0.61 \\
        \texttt{AE}  & 4.89 & 3.00 & 5.00 &0.31 \\
         \bottomrule
    \end{tabular}
     \caption{Evaluation Results of MBTI-S2Conv. We evaluate three aspects: \texttt{EI}, \texttt{PS}, and \texttt{AE}, based on ChatGPT, and display the average (Avg.), minimum (Min.), and maximum (Max.) scores, along with the standard deviation (Std.)}
    \label{tab:conv_eval}
\end{table}
\begin{figure*}[!htbp]
  \centering
  \includegraphics[width=0.95\linewidth]{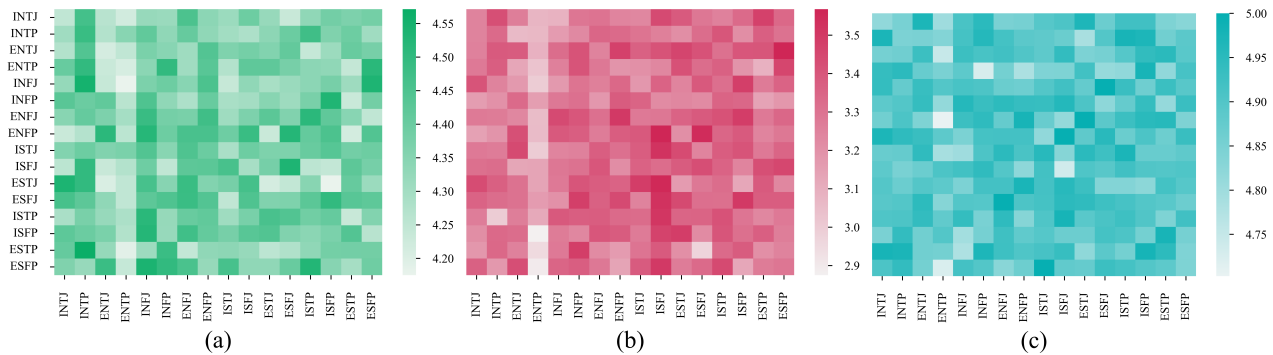}
  \caption{Evaluation scores of MBTI-S2Conv across 16 MBTI personalities, where sub-figures (a), (b) and (c) represent \texttt{EI}, \texttt{PS} and \texttt{AE}. The X-axis and Y-axis is the MBTI personality of the supporter and seeker respectively.}
  \label{eval_mbti_match}
\end{figure*}

Assessing conversations facilitated by LLMs may present limitations in terms of absolute accuracy. Nevertheless, it does offer comparative information. Thus, we employed ChatGPT to assess the quality of generated conversations based on the criteria: emotional improvement (\texttt{EI}), problem solving (\texttt{PS}), and active engagement (\texttt{AE}).

The statistical results of this evaluation are presented in Table~\ref{tab:conv_eval}. The conversations exhibit commendable performance in terms of \texttt{EI} and \texttt{AE}, which indicates that most seekers actively engage in the conversation and experience emotional improvements. However, the efficacy of problem-solving sometimes encounters obstacles attributed to incompatible personalities between the seeker and the supporter.

To explore how interpersonal compatibility affects support conversations, we visualize the assessment scores for different MBTI personality types. As shown in Figure~\ref{eval_mbti_match}, it's clear that the persona significantly influence the  support conversations. 
There are also discoveries that match the conclusions drawn from a study of society~\cite{o2019heuristic}.
For example, in Figure~\ref{eval_mbti_match} (c), a person with the INTP personality shows higher engagement when the supporter has either the ISTP or INTJ personality. This is due to the shared cognitive patterns and personality attributes among INTP, ISTP, and INTJ individuals. 
Their tendency to reflect on their thoughts and analyze situations logically creates a favorable environment for solving problems and communicating smoothly. Their shared value for independence and individual preferences fosters a robust interpersonal bond.

Furthermore, intriguing revelations emerge from our study. Despite the apparent contradiction between INFP and ENTP personalities, with the former leaning towards introversion and the latter embracing extroversion, the complement manifests under certain circumstances.
The compassionate and caring nature of the INFP harmonizes with the rational and analytical tendencies of the ENTP, while the latter's exploratory and adventurous spirit injects novelty and stimulation into the interaction.
Hence, the most suitable supporter personality would be INFP when the seeker exhibits an ENTP personality. 
We also find that ENTPs might not be the eligible supporters due to potential scarcity in empathy and self-centeredness, as indicated by the lighter bands in the three subfigures of Figure~\ref{eval_mbti_match}.

In summation, These insights underscore the pivotal role of interpersonal compatibility in the social support.

\section{Experimental Results of CharacterChat}
\begin{table*}[!htbp]
    \centering
    \small
    \begin{tabular}{c | c c c c c c c  }
    \toprule 
         & BlenderBot-Joint & Vicuna-13b-v1.5 & ChatGPT & ChatGPT$_r$  & ChatGPT$_o$ & CharacterChat$_r$ & CharacterChat$_o$   \\
         \midrule
\texttt{EI}   & 4.16  & 3.98  & 3.81  & 4.42  & 4.39  & 4.39  & \textbf{4.48} \\
\texttt{PS}  & 2.88  & 2.57  & 3.46  & 3.39  & 3.42  & 3.43 &  \textbf{3.50} \\
\texttt{AE}   & 4.82  & 4.62  & 4.61  & 4.87  & \textbf{4.93} & 4.85 & 4.92   \\
         \bottomrule
    \end{tabular}
    \caption{Automatic Evaluation Results. Subscript $r$ indicates the random supporter, and $o$ indicates the optimal supporter chosen by our interpersonal matching model as the profile. Model without subscripts refer to the absence of the profile.}
    \label{tab:auto_eval}
\end{table*}

\subsection{Automatic Evaluation}

Based on the criteria \texttt{EI}, \texttt{PS}, and \texttt{AE}, we compare our CharacterChat with following using 256 simulated seekers in MBTI-1024 and evaluate them by ChatGPT:

\begin{itemize}
    \item \textbf{BlenderBot-Joint~\cite{liu2021towards}.} A pretrained emotional support conversation model that can employ strategies to control the dialogue process.
    \item \textbf{Vicuna-13b-v1.5~\cite{vicuna2023}.} A ChatGPT-like model developed with 13 billion parameters, which is considered one of the most powerful open-source LLMs. We used the latest version from Hugging Face\footnote{https://huggingface.co/lmsys/vicuna-13b-v1.5}.
    \item \textbf{ChatGPT~\cite{ChatGPT}.} An advanced large language model developed by OpenAI. The API version used in this work is \texttt{gpt-3.5-turbo-16k-0613}.
\end{itemize}

As depicted in Table~\ref{tab:auto_eval}, the following conclusions can be drawn from the data analysis: (1) A comparison between BlenderBot-Joint and CharacterChat underscores the unsuitability of emotional support conversation methods for social support interactions. (2) The competence of LLM assistants in handling seekers with diverse personas is notably challenged in the absence of the adapted supporter profile. This inadequacy is apparent in the performance of ChatGPT and Vicuna-13b-v1.5.(3) Within the domain of social support conversations, the specialized CharacterChat exhibits superior performance compared to the more general ChatGPT. This distinction becomes particularly evident when evaluating CharacterChat$_o$ in contrast to ChatGPT$_o$. (4) The interpersonal matching mechanism, which dispatches an optimal supporter based on the seeker's persona greatly improves the support effect. This advancement is discernible through a comparison between CharacterChat$_o$ and CharacterChat$_r$.

The above findings highlight the clear effectiveness of CharacterChat in providing social support and gaining significant advantages from interpersonal matching.

\subsection{Human Evaluation}
\begin{table}[!htbp]
    \centering
    \small
    \begin{tabular}{c | c c c c c c  }
    \toprule 
         & S1	& S2	&S3 &S4	&S5 &Average \\
         \midrule
\texttt{EI}    & -0.078  & 0.261  & 0.265  & 0.561  & 0.223  & 0.246  \\
\texttt{PS}    & 0.116  & 0.261  & 0.212  & 0.601  & -0.048  & 0.228  \\
\texttt{AE}    & -0.034  & 0.526  & 0.197  & 0.411  & -0.406  & 0.138  \\
         \bottomrule
    \end{tabular}
    \caption{Pearson Correlation between seeker-supporter compatibility and the seeker's rate score on the criteria \texttt{EI}, \texttt{PS} and \texttt{AE}. S1-S5 are the human seekers.}
    \label{tab:pearson}
\end{table}
For deeper understanding of the effectiveness of interpersonal matching in personal social support conversations, we invite 5 volunteers to act as seekers and engage in conversations with 16 virtual supporters from the MBTI-1024 Bank. Each supporter possesses a distinct MBTI personality type. The seekers were then tasked with evaluating their experiences based on the criteria \texttt{EI}, \texttt{PS} and \texttt{AE} as above.

We used the Pearson correlation coefficient to measure the relevance between seeker-supporter compatibility and seeker ratings on specific criteria. The result is displayed in Table~\ref{tab:pearson}. While the Pearson correlations may be not a consistently positive number for every individual seeker, the comprehensive results demonstrate an overall positive correlation. This suggests that the degree of interpersonal matching significantly influences seeker' experiences in a manner that is positively aligned. Notably, the aspect of emotional improvement (\texttt{EI}) exhibited the most pronounced correlation, thus affirming the pivotal role of interpersonal matching.

% We employed the Pearson correlation coefficient to measure the relationship between the seeker-supporter compatibility computed by our interpersonal matching model and the seekers' ratings according to the specified criteria,. The result is displayed in Table~\ref{tab:pearson}. While the Pearson correlations may not consistently exceed 0 for every individual seeker, the comprehensive results demonstrate an overall positive correlation. This suggests that the degree of interpersonal matching significantly influences seeker' experiences in a manner that is positively aligned. Notably, the aspect of emotional improvement (\texttt{EI}) exhibited the most pronounced correlation, thus affirming the pivotal role of interpersonal matching.

% We use Pearson correlation to evaluate the relevance between the seeker-supporter compatibility scored by our interpersonal matching model and the seeker's rate score on these criteria. The result is presented in Table~\ref{tab:pearson}. Nonetheless that the Pearson correlations are not always large than 0 for specific seekers, the comprehensive result shows the positive correlation, indicating that the degree of interpersonal matching does indeed affect users' experiences in a positive correlated manner in social support conversations. Specifically, the emotional improvement (\texttt{EI}) of the seeker shows the strongest correlation, verifying the importance of interpersonal match. 

\section{Conclusion}
In this work, we have introduced the \textbf{S}ocial \textbf{S}upport \textbf{Conv}ersation (S2Conv) framework as a novel solution to the challenges faced by traditional emotional support methods. Our central contribution involves the creation of the MBTI-1024 Bank, housing a diverse array of virtual characters with distinct profiles, and MBTI-S2Conv, the social support conversations between the characters in the MBTI-1024 Bank. Both of them facilitate the development of CharacterChat, the first S2Conv system which encompasses a persona- and memory-based conversational model, and an interpersonal matching plugin model to dispatch the compatible supporter for the seeker with specific persona. 
Our work not only highlights the remarkable capabilities of CharacterChat for personalized social support but also emphasizes the pivotal role of interpersonal matching for in enhancing the effect of support. We believe that this study could rise more attention on interpersonal matching mechanism for goal-oriented human-machine dialogue in the futures.

% Our work not only highlights the remarkable capabilities of CharacterChat in providing personalized social support but also underscores the pivotal role of interpersonal matching in enhancing the effectiveness of support. We believe that this study could attract more attention to the importance of interpersonal matching mechanisms in goal-oriented human-machine dialogues in the future

\newpage
\bibliography{aaai24}

\appendix
\section{Appendix A. Information about MBTI}
\label{sec:mbti}
\begin{figure*}[!h]
    \centering
    \includegraphics[width=0.99\linewidth]{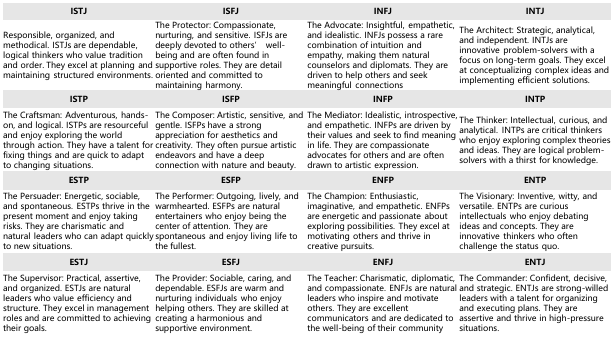}
    \caption{The description for 16 MBTI personalities.}
    \label{fig:prompt-up}
\end{figure*}

The Myers-Briggs Type Indicator (MBTI) is a widely used personality assessment tool based on the theory proposed by Carl Jung. It categorizes individuals into different personality types based on four key dimensions:
\begin{itemize}
    \item Extraversion (E) vs. Introversion (I) - Refers to how individuals gain energy: through external stimuli and interactions (extraversion) or by being more reserved and reflective (introversion).
    \item Sensing (S) vs. Intuition (N) - Describes how individuals gather information: either through concrete details and practicality (sensing) or through patterns and possibilities (intuition).
    \item Thinking (T) vs. Feeling (F) - Represents how individuals make decisions: based on objective analysis and logic (thinking) or by considering personal values and emotions (feeling).
    \item Judging (J) vs. Perceiving (P) - Reflects how individuals approach the outside world: with a structured and organized lifestyle (judging) or with a more flexible and adaptable approach (perceiving).
\end{itemize}
By combining the preferences from each dimension, individuals are assigned a four-letter type that represents their unique personality profile, such as `INTJ', `ESFP', or `ENFJ'. The description of each personality is as follows:

The MBTI is often used in personal development, career counseling, and team-building exercises to gain insights into individual behavior and communication styles. However, it's essential to understand that the MBTI has both supporters and critics, and its scientific validity has been a subject of debate within the field of psychology. 

\end{document}